\pdfoutput=1

\documentclass[11pt]{article}

\usepackage[]{EMNLP2022}

\usepackage{times}
\usepackage{latexsym}
\usepackage{booktabs}

\usepackage{amsmath}
\usepackage{amssymb}
\usepackage{makecell} 
\newcommand{\tabincell}[2]{\begin{tabular}{@{}#1@{}}#2\end{tabular}}

\usepackage[T1]{fontenc}

\usepackage[utf8]{inputenc}

\usepackage{microtype}

\usepackage{inconsolata}

\newcommand{\authorskip}{\hspace{2.5mm}}

\title{
Can Current Explainability Help Provide References in Clinical Notes to Support Humans Annotate Medical Codes?
}

\author{
 Byung-Hak Kim\(^{\dagger,1}\) \authorskip Zhongfen Deng\(^{2}\) \authorskip Philip S. Yu\(^{2}\) \authorskip
 Varun Ganapathi\(^{1}\) \\[2mm]
 \small \(^{\dagger}\)project lead \\[2mm]
 \(^{1}\)AKASA, Inc. \authorskip \(^{2}\)University of Illinois Chicago\\
}

\begin{document}
\maketitle
\begin{abstract}
The medical codes prediction problem from clinical notes has received substantial interest in the NLP community, and several recent studies have shown the state-of-the-art (SOTA) code prediction results of full-fledged deep learning-based methods. However, most previous SOTA works based on deep learning are still in early stages in terms of providing textual references and explanations of the predicted codes, despite the fact that this level of explainability of the prediction outcomes is critical to gaining trust from professional medical coders. This raises the important question of how well current explainability methods apply to advanced neural network models such as transformers to predict correct codes and present references in clinical notes that support code prediction. First, we present an e\textbf{x}plainable \textbf{R}ead, \textbf{A}ttend, and \textbf{C}ode (xRAC) framework and assess two approaches, attention score-based xRAC-ATTN and model-agnostic knowledge-distillation-based xRAC-KD, through simplified but thorough human-grounded evaluations with SOTA transformer-based model, RAC. We find that the supporting evidence text highlighted by xRAC-ATTN is of higher quality than xRAC-KD whereas xRAC-KD has potential advantages in production deployment scenarios. More importantly, we show for the first time that, given the current state of explainability methodologies, using the SOTA medical codes prediction system still requires the expertise and competencies of professional coders, even though its prediction accuracy is superior to that of human coders. This, we believe, is a very meaningful step toward developing explainable and accurate machine learning systems for fully autonomous medical code prediction from clinical notes.
\end{abstract}

\section{Introduction}
\label{sec:Introduction}
Within current medical systems, the prediction of medical codes from clinical notes is a practical and essential need for every healthcare delivery organization~\citep{Dev21}. A human coder or health care provider scans medical documentation in electronic health records, identifying important information and annotating codes for that specific treatment or service. With a diverse range of medical services and providers (primary care clinics, specialty clinics, emergency departments, mother-baby units, outpatient and inpatient units, etc.), the complexity of human coders' tasks grows, while productivity standards fall as charts take longer to review. Thus, even partial automation of the annotation workflow will save significant time and effort that human coders currently spend. The biggest challenge, however, is directly identifying appropriate medical codes from thousands of high-dimensional codes from unstructured free-text clinical notes~\citep{Dong22}. 

Lately, advanced deep learning-based methods for predicting medical codes based on clinical notes~\citep{Kim21,Sun21,Liu21,Yuan22} have achieved state-of-the-art prediction performance and even reached parity with human coders' performance~\citep{Kim21}. However, most current works on medical code prediction based on deep learning models do not provide the end-user with references from the clinical notes to explain why the predicted codes were presented/chosen. There have been some related works that provide the rationales or text highlights from clinical notes to explain why the predictions were made to support humans clinical decision making~\citep{Taylor21,Cao20,Mullenbach18,Wood22}. However, to the best of our knowledge, there is still a gap in studies that have thoroughly analyzed explainability to extract supporting text for code prediction, especially made by state-of-the-art (SOTA) transformer-based models such as the RAC model~\citep{Kim21}.

Two examples are the attention score-based approach first introduced in~\citet{Mullenbach18} and the model-independent knowlege-distillation based method recently initiated in~\citet{Wood22}. The first approach utilizes the per-label attention mechanism to select key sentences for prediction decisions; however, if the model does not have the per-label attention layer, it cannot generate text snippets, and even worse, applying this method to transformer-based architecture and deploying it to production comes with a range of compute and memory challenges~\citep{Vaswani17}. In the second knowledge-distillation approach, a large neural network is distilled into linear student models in a post-hoc manner without sacrificing much accuracy of the teacher model while retaining many advantages of linear models including explainability and smaller model size, which is beneficial for deployment. 


This paper makes two primary contributions:
\begin{itemize}
\item First, we present a general and explainable xRAC framework that generates evidentiary text snippets for a predicted code, which are oriented towards the needs of a deployment scenario of the RAC model. Then, to better assess the explainability of our xRAC framework, human-grounded evaluations is conducted with two groups of internal annotators, one group with and one group without medical coding expertise. We find that the proposed xRAC framework can benefit professional coders but not lay annotators who lack relevant expertise and competencies.


\item Second, we propose code-prior matching and text-prior matching losses to augment the original binary cross-entropy (BCE) loss used to train the RAC model. Because trained models with BCE loss typically tend to focus more on the frequent medical codes and their associated clinical notes portions of the dataset, these new losses are to help distribute the gradient update evenly across all of the codes and note tokens, regardless of code's frequency and token's relevance to codes, so as to improve the xRAC model's prediction as a whole. 
\end{itemize}
\section{xRAC Framework}
\label{sec:RAC-X}


\subsection{xRAC-ATTN}
\label{ssec:xRAC}
The original RAC architecture is built on the code-title guided attention module that considerably improves the per-label attention mechanism first introduced in \citet{Mullenbach18}. This enhanced attention module is to address the extreme sparsity of the large code output space with so called code-title embedding. Because code titles (or descriptions) contain important semantic information and meaning of the codes, the RAC model obtains its embedding from its textual description as shown in the Table \ref{table: humanevalexamples} examples. Specifically, the code description is fed into an embedding layer, which is then followed by a CNN and Global Max Pooling layer to learn the code embedding. 

Therefore, the first xRAC-ATTN directly leverages the attention scores learned in the RAC model to generate the evidence text for each code \(i\). In particular, the attention scores \(\textbf{w}_{i}^\text{ATTN}=(w_{i,1}^{\text{ATTN}}, ...,w_{i,n_{x}}^{\text{ATTN}})\) on the input tokens for code-\(i\) is computed as follows:
\begin{equation}\label{eq:racattn}
\begin{split}
    \textbf{w}_{i}^{\text{ATTN}} = \text{Softmax}\left(\frac{\mathbf{e_i} \mathbf{U^{T}_{x}}}{\sqrt{d}}\right),
\end{split}    
\end{equation}
where \(\mathbf{e_{i}}\in \mathbb{R}^{1\times d}\) is one row of \(\mathbf{E_t} \in \mathbb{R}^{n_{y}\times d}\) which is the code embeddings from the code descriptions, \(\mathbf{U_x} \in \mathbb{R}^{n_{x}\times d}\) is the text representation outputted by the Reader, \(d\) is the dimension of code embedding, \(n_x\) is the number of tokens in the input document, and \(n_y\) is the number of codes in the dataset.

\subsection{xRAC-KD}
\label{ssec:xRAC-KD}
The application of the original idea of knowledge distillation~\citep{Hinton14} requires specific adjustments to the problem setting of medical codes prediction. Knowledge distillation is typically used to train a compact neural network from a large or ensemble of neural network models. Unlike those standard approaches, xRAC-KD transfers the large RAC-based ``teacher'' model into a set of reliable and explainable ``student'' linear models by distilling the predictions made by the large teacher model. 

Assume that we have a trained ``teacher'' neural network \(f_{\text{teacher}}(\textbf{x}_{t})\) and training data.\footnote{Note that because there is flexibility in using different representations for the same clinical note, we use different notations \(\textbf{x}_{t}\) and \(\textbf{x}_{s}\) to denote a tokenized clinical note. We use Word2Vec for \(\textbf{x}_{t}\) and bag of words for \(\textbf{x}_{s}\).} xRAC-KD approximates \(f_{\text{teacher}}(\textbf{x}_{t})\) with a collection of student linear models \(f_{\text{student}}(\textbf{x}_{s})=(f_{s,0}(\textbf{x}_{s}), ..., f_{s,n_{y}}(\textbf{x}_{s}))\) defined as
\begin{equation}
\begin{split}
    f_{s,i}(\textbf{x}_{s}) = \textbf{w}_{i}^{\text{KD}}\textbf{x}_{s},
\end{split}    
\end{equation}
where \(\textbf{w}_{i}^\text{KD}=(w_{i,1}^{\text{KD}}, ...,w_{i,n_{x}}^{\text{KD}})\). \(f_{\text{teacher}}(\textbf{x}_{t})\) produces predicted probability vector \(\hat{\textbf{y}}_{t}=(\hat{y}_{t,1}, ...,\hat{y}_{t,n_{y}})\). First xRAC-KD converts \(\hat{\textbf{y}}_{t}\) to \(\textbf{q}_{t}=(q_{t,1}, ..., q_{t,n_{y}})\) which is defined as  
\begin{equation}\label{eq:logit}
\begin{split}
    q_{t,i} = T\text{logit}(\hat{y}_{t,i})=T\log\bigg(\frac{\hat{y}_{t,i}}{1-\hat{y}_{t,i}}\bigg),
\end{split}    
\end{equation}
where a temperature parameter \(T\) is to adjust the logit values and set it to 1 for convenience. 

Then, as a distillation loss to train the student models \(f_{\text{student}}(\textbf{x}_{s})\), xRAC-KD uses the L1 regularized regression loss between \(\textbf{q}_{t}\) and the student's predicted output vectors \(\textbf{q}_{s}\)  written as follows 
\begin{equation}
\begin{split}
     ||\textbf{q}_{t}-\textbf{q}_{s}||_{2} + \lambda||\textbf{w}_{i}^{\text{KD}}||_{1},
\end{split}    
\end{equation}
with \(\lambda\) parameter. xRAC-KD does not use any additional loss term with respect to the training data's hard labels (either 0 or 1).
Once the distilled student models \(f_{\text{student}}(\textbf{x}_{s})\) are ready, xRAC-KD finally transforms the output vector \(\textbf{q}_{s}\) back to the prediction vector \(\hat{\textbf{y}}_{s}=(\hat{y}_{s,0}, ...,\hat{y}_{s,n_{y}})\) easily as follows:
\begin{equation}\label{eq:expit}
\begin{split}
    \hat{y}_{s,i} = \text{expit}\bigg(\frac{{q}_{s,i}}{T}\bigg)=\frac{1}{1+\exp(-{q}_{s,i}/{T})}.
\end{split}    
\end{equation}

The logit and expit transforms defined in Eq. (\ref{eq:logit}) and (\ref{eq:expit}) pairs that are inverse to each other are a fundamental improvement over the initial method presented in \citet{Wood22}. Previously, the distilled models showed consistently low precision scores and it was hypothesized for the independence of the distilled linear models. However, by comparing the first and last rows of Table \ref{table: modelperformance}, it turns out that this new pair has resulted in a clear outperformance across the board over the logistic regression baseline unlike the initial approach.

\subsection{Supporting Text Extraction}
\label{ssec:extract}
Lastly, the evidence text of the xRAC-ATTN and xRAC-KD models is constructed by first locating the \emph{n}-gram with the highest average weight score for each code \(i\) calculated as
\begin{equation}
\begin{split}
    \arg\max_{j}\sum_{n-gram}w_{i,j},
\end{split}    
\end{equation}
then \emph{m} tokens on either side of the \emph{n} gram are included to obtain the final subsequence of evidence with length of \(n+2m\). We set \(n\) to 4 and \(m\) to 5.

\subsection{xRAC with Augmented Losses}
\label{ssec:uRAC}

The RAC model utilizes a transformer encoder and an attention-based architecture to attain SOTA performance. It also makes use of code descriptions to obtain code embeddings. Although the code embeddings obtained from the code description capture the semantic meaning of each code, due to the natural characteristics of medical coding, most of the codes appear just a few times compared to other common codes associated with common diseases. 

Similarly, not all tokens in a given piece of text can be learned sufficiently and equally during the training process; therefore, frequent code embedding (as well as token embeddings) will receive more updates than infrequent codes (and tokens). In other words, trained models with BCE loss tend to focus more on the frequent codes and their associated clinical notes portions in the dataset; therefore, we propose code-prior matching and text-prior matching losses to supplement the BCE loss to encourage the models better handle imbalance issues and improve the model's overall prediction. 

\textbf{Code Prior Matching (CPM):} To alleviate the issue of frequent codes receiving more updates than infrequent codes during training, CPM is applied to the second to the last output of the Coder, \(\mathbf{V_{x}} \in \mathbb{R}^{n_{y}\times d}\) defined as  

\begin{equation}
\label{eq:vx}
\begin{split}
    \mathbf{V_{x}} = \text{Softmax}\left(\frac{\mathbf{E_{t}}\mathbf{U^{T}_{x}}}{\sqrt{d}}\right)\mathbf{U_{x}}.
\end{split}
\end{equation}

The CPM can help the model learn evenly across all codes, regardless of frequency, by imposing constraints on the learned \(\mathbf{V_x}\). This prior matching module is implemented by a discriminator \(D_{\text{cpm}}\), which shares the same structure as \(D_{\text{lpm}}\) in \citet{Deng21} and introduces a regularization loss for each code as
\begin{equation}\label{eq:prloss_foreachcode}
\begin{split}
    l_{c}^{i} = -(\mathbb{E}_{{\mathbf{c_p}}\sim\mathbb{Q}}[\log D_{\text{cpm}}(\mathbf{c_p})] + \\ \mathbb{E}_{\mathbf{v_i}\sim\mathbb{P}}[\log(1-D_{\text{cpm}}(\mathbf{v_i}))]),
\end{split}
\end{equation}
where $\mathbf{v_i} \in \mathbb{R}^{1\times d}$ is one row of \(\mathbf{V_x}\) which is the vector for one code in the dataset, $\mathbb{P}$ is the code embedding distribution learned by the model, $\mathbf{c_p}$ is a prior vector of the same size as \(\mathbf{v_i}\) for the given code generated by a uniform distribution $\mathbb{Q}$ in the interval of $[0, 1)$, and \(l_c^{i}\) is the prior matching loss for code-\(i\).\footnote{We chose a compact uniform distribution on \([0, 1)\) as the prior, which worked better in practice than other priors, such as Gaussian, unit ball, or unit sphere as shown in previous works~\citep{Deng21,Devon19}.} We take the average of \(l_c^{i}\) losses from all codes to obtain the final CPM loss \(L_{\text{C}}\) as follows:
\begin{equation}\label{eq:code_priormatchingloss}
    L_{\text{C}} = \frac{1}{n_y} \sum_{i=1}^{n_y} l_{c}^{i}.
\end{equation}

\textbf{Text Prior Matching (TPM):} In the RAC model, not all tokens in a particular clinic text note can be learned equally, as the Reader focuses more on tokens related to frequent codes in the data set. To help the model's gradient be updated equally for all tokens in the input, a TPM loss is applied on \(\mathbf{U_{x}}\) output of the Reader. The TPM is also implemented by a discriminator \(D_{\text{tpm}}\) similar to \(D_{\text{cpm}}\), where it introduces another prior matching loss $L_{\text{T}}$ shown as
\begin{equation}\label{eq:tpmloss_foreachcode}
\begin{split}
    l_{t}^{i} = -(\mathbb{E}_{{\mathbf{t_p}}\sim\mathbb{Q}}[\log D_{tpm}(\mathbf{t_p})] +\\ \mathbb{E}_{\mathbf{u_i}\sim\mathbb{P}}[\log(1-D_{tpm}(\mathbf{u_i}))]),
\end{split}
\end{equation}
\begin{equation}\label{eq:text_priormatchingloss}
    L_{\text{T}} = \frac{1}{n_x} \sum_{i=1}^{n_x} l_{t}^{i},
\end{equation}
where \(\mathbf{u_i}\) is one row of $U_x$ that is the embedding of a token in the input document, $\mathbb{P}$ is the distribution of text embedding learned by the model, $\mathbf{t_p}$ is the prior embedding vector for the given token in the input document also generated by a uniform distribution $\mathbb{Q}$ in the interval of $[0,1)$, and \(l_t^{i}\) is the TPM for a token in the input; we then use the average loss for all tokens in the input document as the final TPM loss \(L_\text{T}\), similar to Eq. (\ref{eq:code_priormatchingloss}). This loss can make the model evenly learn the embeddings for all tokens in the input, which will be fed to the Coder for code prediction. 

\textbf{Overall Training Loss:} Finally, the total augmented loss is written as
\begin{equation}\label{eq:total_loss}
    L_{\text{total}} = L_{\text{BCE}} + \alpha * L_{\text{C}} + \beta*L_{\text{T}},
\end{equation}
where \(\alpha\) and \(\beta\) are parameters to balance \(L_{\text{C}}\) and \(L_{\text{T}}\) respectively. The updated RAC model trained with \(L_{\text{total}}\) instead of BCE loss, is first used for xRAC-ATTN and its performance is shown in the third row of Table \ref{table: modelperformance}. Although we used the original RAC model as a teacher model to distill from in xRAC-KD, this updated RAC model can also be used. Comparing the second (RAC model trained with BCE loss) and third rows (updated RAC model trained with \(L_{\text{total}}\)) in Table \ref{table: modelperformance} shows modest improvements in both standard and hierarchical micro F1 scores, indicating that prior matching modules modestly help to address the imbalanced issues.

\begin{table*}[ht]
    \caption{Medical codes prediction results (in \%) by ML systems on the MIMIC-III-full-label testing set as described in \citet{Kim21}. The bold value shows the best (and highest) value for each column metric. The logistic regression results are taken from \citet{Mullenbach18}, and the RAC results come from \citet{Kim21}. All numbers are the results of a single run with fixed random seeds, as practiced in the previous literature~\citep{Kim21,Mullenbach18} for apples-to-apples comparisons. Note that our baseline is the most recent SOTA model RAC, and our xRAC-ATTN outperforms RAC in most metrics.}. 
    \label{table: modelperformance}
    \centering
    \begin{tabular}[t]{l|l|l|l|l}
    \toprule[\heavyrulewidth]
    \textbf{Model} & \textbf{AUC} & \textbf{Standard F1} & \textbf{Precision@n} & \textbf{Hierarchical F1} \\
    & Macro Micro & Macro Micro & 5 \ \ \ \ \ \ \ 8 \ \ \ \ \ \ \ 15 & CoPHE Set-Based \\
    \midrule
    Logistic Regression & 56.1 \ \ \ \ 93.7 & \ \ 1.1 \ \ \ 27.2 & \ \ \ \ \ \ \ \ \ \ 54.2 \ \ 41.1 \\  
    RAC & \textbf{94.8} \ \ \ \ \textbf{99.2} & \textbf{12.7} \ \ \ 58.6 & \textbf{82.9} \ \ 75.4 \ \ \textbf{60.1} & 62.7 \ \ \ \ \ 64.0 \\ 
    \midrule
    xRAC-ATTN (ours) & \textbf{94.8} \ \ \ \ 99.1 & 12.6 \ \ \ \textbf{58.8} & \textbf{82.9} \ \ \textbf{75.6} \ \ \textbf{60.1} & \textbf{62.9} \ \ \ \ \ \textbf{64.3} \\
    xRAC-KD (ours) & 93.6 \ \ \ \ 98.7 & \ \ 7.4 \ \ \ 46.0 & 69.4 \ \ 61.6 \ \ 48.6 & 51.8 \ \ \ \ \ 54.5 \\ 
    \bottomrule[\heavyrulewidth]
    \end{tabular}
\end{table*}
\section{Experimental Results}
\subsection{MIMIC-III Dataset}
\label{app:dataset}
The MIMIC-III Dataset (MIMIC v1.4~\citet{Johnson16}) is a freely accessible medical database that contains de-identified medical data from over 40,000 patients who visited the Beth Israel Deaconess Medical Center between 2001 and 2012.\footnote{One reason for using the MIMIC-III dataset for this study is that it has been used as standard benchmark in previous studies~\citep{Kim21,Mullenbach18}, allowing meaningful head-to-head comparisons with our work. We believe that the proposed xRAC model is not limited to the MIMIC-III dataset and will also work well with a MIMIC-IV dataset, but MIMIC-IV-Note is currently not available to the public.} We extract the discharge summaries and the corresponding medical codes, for this study. For a direct comparison with previous works, we use the same data processing, and data split described in \citep{Mullenbach18}. This processing results in  47,724 samples for training, 1,632 and 3,373 samples for validation and testing, respectively, with an average number of 16 codes assigned to each discharge summary. More dataset statistics, can be found in Table 2 of \citep{Mullenbach18}.  

\subsection{Training Details}
\label{app:modeltraining}
The xRAC models follow the same training details as the RAC model, which can be found in the original RAC paper~\cite{Kim21}. The xRAC-ATTN model is also trained with the same hyperparameters as the RAC model.\footnote{The maximum sequence length is 4096, and there are four stacks of attention layers with single attention head. The code and text embedding dimensions are 300 and the batch size is 16.} The xRAC-ATTN model's extra hyperparameters include $\alpha$ and $\beta$ in Eq. (\ref{eq:total_loss}), with values of 0.5 and 0.8 respectively. The temperature for the xRAC-KD model is set to 1, \(\lambda\) to 1e-3, and the maximum iteration for the training is set to 800.\footnote{For the choices of hyper-parameters, we fine-tuned the model by running a linear search of these hyper-parameters to find the best value at which the model's performance peaks.}

\subsection{xRAC Model Performance}
In addition to the same standard flat metrics used in previous RAC model evaluations, recently introduced hierarchical metrics (e.g. CoPHE~\citep{Falis21}, set-based metrics~\citep{Kosmopoulos15}) are used. These two metrics take the hierarchical structure of the ICD codes tree into consideration for evaluating codes prediction. The CoPHE metric further utilizes depth-based hierarchical representation and the count of codes at different ancestral levels of the tree to evaluate model's prediction, providing more meaningful evaluation in this context. 

The results of the xRAC-ATTN and the xRAC-KD are shown in the last two rows of Table \ref{table: modelperformance} respectively. First, when compared to the prior RAC model trained with BCE loss, the xRAC-ATTN model improves both standard and hierarchical micro F1 scores, as noted by comparing the second and third rows of Table \ref{table: modelperformance}, suggesting that the prior matching modules modestly help and effectively improve the SOTA scores. Second, while the xRAC-KD student model (shown in the last row) performs slightly worse than that of the RAC-based teacher model (shown in the second row), it still significantly outperforms the logistic regression baseline (shown in the first row, which was trained from scratch and has the same level of model complexity) across the board, which was not the case in \citet{Wood22}.

\begin{table*}
    \caption{Two example questions provided for human evaluation: The codes in these two questions are the same, ``521.00, Dental caries, unspecified", however, the two explanation text snippets classified as A) and B) are extracted by two different models, xRAC-ATTN and xRAC-KD. The information about the models is hidden from human annotators, and the order of text snippets for the same code is permuted to prevent the annotators from guessing the models based on the order. Note that \textbf{HI}, \textbf{I}, and \textbf{IR} stand for Highly Informative, Informative, and Irrelevant, respectively.}
    \label{table: humanevalexamples}
    \centering
    \begin{tabular}{p{1.5cm}<{\centering}|p{3.3cm}<{\centering}|p{7.5cm}<{\centering}|p{.4cm}<{\centering}|p{.4cm}<{\centering}|p{.4cm}<{\centering}}
    \toprule[\heavyrulewidth]
    \textbf{Question ID} & \textbf{Code and Description} & \textbf{Explanation Text Snippet} & \textbf{HI} & \textbf{I} & \textbf{IR} \\
    \midrule
    1 & \tabincell{c}{521.00, Dental caries, \\unspecified} & \tabincell{c}{A) surgical or invasive procedure left \\ **and right heart catheterization** \\ coronary angiogram multiple dental extractions} & & \\ 
    \midrule
    1 & \tabincell{c}{521.00, Dental caries, \\unspecified} & \tabincell{c}{B) balloon s p dental extractions \\ **s p exploratory laparotomy** \\ and cholecystectomy fungal sepsis discharge} & & \\ 
    \bottomrule[\heavyrulewidth]
    \end{tabular}
\end{table*}

\begin{table*}
    \caption{The overall informativeness of xRAC-ATTN and xRAC-KD retrieved explanatory text snippets. The left half represents the outcome of Group A's annotation, while the right half represents the outcome of Group B's evaluation. \textbf{HI}, \textbf{I}, and \textbf{IR} stand for Highly Informative, Informative, and Irrelevant, respectively. Percent denotes the ratio of informative text snippets (HI and I) to the total extracted snippets, which is 3,813 (in \%).}
    \label{table:heval-overall}
    \centering
    \begin{tabular}[t]{l|l|l}
    \toprule[\heavyrulewidth]
    \textbf{Model} & \textbf{Group A (Lay Annotators)} & \textbf{Group B (Professional Coders)} \\
    & HI \ \ \ \ \ \ \ I \ \ \ \ \ \ \ \ \ IR \ \ \ \ \ \ \ \ Percent & \ HI \ \ \ \ \ \ I \ \ \ \ \ \ \ \ \ IR \ \ \ \ \ \ \ \ \ Percent \\
    \midrule
    xRAC-ATTN & 1652 \ \ \ 1389 \ \ \ \ \ 772 \ \ \ \ 79.75 & \ 1283 \ \ 1094 \ \ \ 1436 \ \ \ \ \ 62.34 \\ 
    xRAC-KD & \ \ 865 \ \ \ 1318 \ \ \ 1630 \ \ \ \ 57.25 & \ \ \ 145 \ \ \ \ 212 \ \ \ 3456 \ \ \ \ \ \ \ 9.36 \\ 
    \bottomrule[\heavyrulewidth]
    \end{tabular}
\end{table*}

\begin{table*}
    \caption{The evaluation agreements on Highly Informative and Informative text snippets between Groups A and B as measured by Jaccard Similarity (in \%). Note that we evaluated the annotation consistency between two groups as described in Section~\ref{sec:Explainability Human Evaluation}, and the annotation consistency (or correctness) of lay annotators (Group A) is lower than 40\% even provided with the same textual references as for professional coders (Group B).}
    \label{table:heval-buckets}
    \centering
    \begin{tabular}[t]{l|l}
    \toprule[\heavyrulewidth]
    \textbf{Model} & \textbf{Jaccard Similarty} \\
    & HI \ \ \ \ \ \ \ \ \ I \\
    \midrule
    xRAC-ATTN & 39.2 \ \ \ \ \ \ 18.5 \\ 
    xRAC-KD & \ \ 7.0 \ \ \ \ \ \ \ \ 5.0 \\ 
    \bottomrule[\heavyrulewidth]
    \end{tabular}
\end{table*}

\subsection{Human-Grounded Evaluation}
\label{sec:Explainability Human Evaluation}
\textbf{Human Evaluation Design:} 
Human-grounded evaluation is important for evaluating the explainability. Because medical code annotation involves domain knowledge specific to  medical coding, human evaluation is challenging; thus, we conducted a human evaluation with two groups of internal annotators. \textbf{Group A} had two annotators without medical coding experience and \textbf{Group B} had six certified professional coders. While both groups followed the same annotation instructions and guidelines, Group A was supervised by one manager and Group B was supervised by two managers with professional coder management experience to ensure annotation consistency (i.e., inter-annotator agreement) within each group. Group A worked full-time for two weeks to finish all the annotation, while Group B worked part-time for three weeks. Because the two groups of annotators involved in the human evaluation process are well aware that the task involves the medical notes of anonymized patients, the study does not require IRB approval and does not raise any ethical concerns. 


\textbf{Annotation Task Design:} We select the overlap of codes predicted between the xRAC-ATTN and xRAC-KD models on the MIMIC-III-full-label testing set and combine the code descriptions and the corresponding textual explanations generated by each model together in a question sheet~\footnote{The MIMIC-III dataset's entire test set is used for human evaluation. Specifically, both the xRAC-ATTN and xRAC-KD models take clinical note from each example in the test set as input and predict multiple codes associated with this note. Because each model can predict differently for each example in the test set, we select all the test examples from the two models that are predicted with the same codes to compare their explainability. As a result, there are a total of 3,813 test examples predicted with the same codes by the xRAC-ATTN and xRAC-KD models.}. We then provide the sheet to Groups A and B for evaluation. Specifically, the question sheet contains six columns which are Question ID, Code and Description, Explanation Text Snippet, Highly Informative, Informative, and Irrelevant (see Table~\ref{table: humanevalexamples} for sample questions). Each code has two different text snippets extracted by two models, respectively. The annotators need to assign one of the three choices, which are highly informative, informative, and irrelevant to every explanation text snippet extracted to support the appearance of the predicted code. 

Highly informative is defined as if the text snippet provides an accurate explanation for the predicted code. Otherwise, it is informative as long as the annotators believe that the text snippet adequately explains the presence of the given code, is related to the code's description, or has a close meaning to the code's description. Because the medical note contains domain knowledge, it is difficult for annotators to assign a finer-grained scale to the textual evidence when deciding between highly informative, informative, and irrelevant. 

The final question sheet has a total of 3,813 codes predicted with different supporting text snippets. Unlike all previous studies, which typically collect less than 100 samples from clinicians (e.g., \citet{Mullenbach18}), the task design of our study is quite unique, as is the volume of questions to our knowledge.

\textbf{Human Evaluation Results:} The results of human evaluation for the explainability of xRAC framework are shown in Tables~\ref{table:heval-overall} and \ref{table:heval-buckets}. Table \ref{table:heval-overall} shows the overall result of the informativeness of the text snippets extracted by xRAC-ATTN and xRAC-KD. The percentage column in Table \ref{table:heval-overall} represents the percentage of explanations annotated as highly informative or informative, excluding irrelevant explanations. Thus, the irrelevant explanations generated by our model are about 20-40\% as shown in Table \ref{table:heval-overall}. 

One can see that there is a much larger gap in xRAC-KD between Group A and Group B than between xRAC-ATTN. Each group of annotators adhered to use the same standard to evaluate the textual explanation and was monitored by managers with professional coder management experience to ensure that there was no annotation variation among annotators in the same group. However, the large deviation between the two groups (Groups A and B) is understandable due to the domain knowledge gap between professional coders and lay annotators. Because of their limited medical knowledge and understanding, lay annotators tend to assign more highly informative and informative to the extracted textual explanation. Whereas, professional coders are much stricter on the informativeness of textual explanations. 

In other words, this implies that xRAC-ATTN is a more viable choice than xRAC-KD to extract a text snippet from clinical notes to support code prediction. However, Table~\ref{table:heval-buckets} shows that the consistency score measured by Jaccard Similarity between two groups is lower than 40\% even with xRAC-ATTN. This suggests that the automated extraction system must continue to rely on professional coders' feedback and domain experience, and that text snippets alone are insufficient to replace them. In other words, there is still room to improve explainability for a lay person without expertise to appropriately code.  



\section{Conclusion} 
\label{sec:Discussion}
In this paper, a xRAC framework is presented to obtain supporting evidence text from clinical notes that justify the predicted medical codes from medical code prediction systems. We have demonstrated that the proposed xRAC framework may help even complex transformer-based models (e.g., RAC model) to attain high accuracy with a decent level of explainability (which is of high value for deployment scenarios) through quantitative experimental studies and qualitative human-grounded evaluations. It was also shown for the first time that, given the current state of explainability methodologies, using the proposed explainable yet accurate medical codes prediction system still requires professional coders' expertise and competencies.
\section*{Limitations}
The current human-grounded evaluation studies only a simplified scenario: the impact of clinical-text-based explanations provided alongside predictions on explainability as judged by humans with and without professional coding backgrounds. This exercise sheds light on a key element that is necessary for these AI coding-based models to be useful in real-world deployment scenarios, but does not definitively ascertain that these coding predictions provided alongside explanations of the prediction would enable a transition to AI-driven coding autonomously. First, we have not studied how to incorporate the proposed xRAC framework into a human-in-the-loop situation with human coder feedback, which may be a very common scenario of deployment in practice. Second, we have not compared a full AI-driven coding model with humans-in-the-loop to a human-only process, in terms of speed, manpower needed, and accuracy. Limitations of the prediction model may become relevant in these situations, as human coders must occasionally combine disparate pieces of information together~\citep{Dong22}. Third, while the MIMIC-III dataset provides a useful benchmark for evaluating approaches, it is not representative of the wide range of clinical notes, so it would be beneficial to expand to other data sets with a wider range of codes.


\section*{Ethics Statement}
First and foremost, an automated and explainable machine learning system for medical code prediction aims to streamline the medical coding workflow, reduce the backlog of human coders by increasing productivity, and assist human coders quickly navigating complex and extended charts while reducing coding errors~\citep{Crawford13}. Second, an automated and explainable system is designed to lessen the administrative burden on providers, allowing them to focus on providing care rather than mastering the complexities of coding. Furthermore, better automated and explainable software can improve clinical documentation, enhance the overall picture of its quality, and eventually redirect lost healthcare dollars to more meaningful purposes~\citep{Shrank19}.

\section*{Acknowledgements}
The authors would like to thank everyone who helped with the human-grounded evaluation, especially Juliann Chaparro, Tracy Salyers, Peg Leland, Jasmine Porter, Jenn Arlas, Amber Taylor, Stetson Bauman, Colin Wilde, Beth Cable, Dana Hunter, and Amy Raymond. Angela Kilby and Jiaming Zeng's suggestions to improve the manuscripts are also gratefully acknowledged.

\bibliography{anthology,custom}
\bibliographystyle{acl_natbib}

\end{document}